\documentclass[a4paper,fleqn]{cas-sc}
\usepackage[numbers]{natbib}
\usepackage{amsmath}
\usepackage{graphicx}

\DeclareUnicodeCharacter{1D446}{\ensuremath{\mathbb{S}}}
\DeclareUnicodeCharacter{1D441}{\ensuremath{\mathbb{N}}}
\DeclareUnicodeCharacter{1D43F}{\ensuremath{\mathbb{L}}}
\DeclareUnicodeCharacter{1D45B}{\ensuremath{\mathbb{n}}}
\DeclareUnicodeCharacter{03B5}{\ensuremath{\mathbb{ε}}}
\DeclareUnicodeCharacter{2212}{\ensuremath{\mathbb{-}}}
\DeclareUnicodeCharacter{03B1}{\ensuremath{\mathbb{α}}}
\DeclareUnicodeCharacter{03B3}{\ensuremath{\mathbb{γ}}}
\DeclareUnicodeCharacter{226A}{\ensuremath{\mathbb{≪}}}

\def\tsc#1{\csdef{#1}{\textsc{\lowercase{#1}}\xspace}}
\tsc{WGM}
\tsc{QE}
\begin{document}
\let\WriteBookmarks\relax
\def\floatpagepagefraction{1}
\def\textpagefraction{.001}
\let\printorcid\relax % 可去掉页面下方的ORCID(s)

% \shorttitle{<short title of the paper for running head>} 
\shorttitle{Lightweight Design and Optimization methods for DCNNs}    

% \shortauthors{<short author list for running head>}
\shortauthors{Hanhua Long et al.}

%论文标题
\title[mode = title]{Lightweight Design and Optimization methods for DCNNs: Progress and Futures}  

\author[1]{Hanhua Long}
\cormark[1]
\ead{longhanhua@neusoft.edu.cn}
\fnmark[1]
\author[1]{Wenbin Bi}
\ead{biwenbin@neusoft.edu.cn}
\fnmark[2]
\author[1]{Jian Sun}
\ead{sunjian@neusoft.edu.cn}
\fnmark[3]
\address[1]{School of Computer \& Software, Dalian Neusoft University of Information, Dalian 116023, China} %声明第一单位

\cortext[1]{Corresponding author}  %声明通讯作者
\fntext[fn1]{\textbf{Hanhua Long} is with School of Computer \& Software, Dalian Neusoft University of Information, Dalian, LN, China. His research interest focuses on efficient and accelerated deep learning algorithms and systems.}
\fntext[fn2]{\textbf{Wenbin Bi} is with School of Computer \& Software, Dalian Neusoft University of Information, Dalian, LN, China. His research interests include vehicle networking and blockchain.}
\fntext[fn3]{\textbf{Jian Sun} is with School of Computer \& Software, Dalian Neusoft University of Information, Dalian, LN, China. His research interest lie in energy-efficient bio computing and biosignal compression.}

\begin{abstract}
Lightweight design, as a key approach to mitigate disparity between computational requirements of deep learning models and hardware performance, plays a pivotal role in advancing application of deep learning technologies on mobile and embedded devices, alongside rapid development of smart home, telemedicine, and autonomous driving. With its outstanding feature extracting capabilities, Deep Convolutional Neural Networks (DCNNs) have demonstrated superior performance in computer vision tasks. However, high computational costs and large network architectures severely limit the widespread application of DCNNs on resource-constrained hardware platforms such as smartphones, robots, and IoT devices. This paper reviews lightweight design strategies for DCNNs and examines recent research progress in both lightweight architectural design and model compression. Additionally, this paper discusses current limitations in this field of research and propose prospects for future directions, aiming to provide valuable guidance and reflection for lightweight design philosophy on deep neural networks in the field of computer vision.
\end{abstract}


% \begin{highlights}
% \item DCNNs excel in computer vision but challenged by hardware constraints.
% \item Lightweight DCNNs is crucial for mobile and embedded devices.
% \item Review highlighted advances in DCNN lightweight design.
% \end{highlights}

% Keywords
\begin{keywords}
CNN \sep 
Lightweight design \sep 
Model optimization
\end{keywords}

\maketitle

% Main text
\section{Introduction}

With the rapid growth of data volumes and significant improvements in hardware computing capabilities, deep neural networks have experienced rapid development. As the most commonly used and successful network structure in deep neural networks, Deep Convolutional Neural Networks (DCNNs) have taken a dominant position in the field of computer vision. Leveraging their powerful feature extraction capabilities, DCNNs perform exceptionally well in vision tasks such as object detection \cite{Szegedy2013DeepNN, Zhao2018ObjectDW}, image classification \cite{Vailaya2001}, and semantic segmentation \cite{Chen2018DeepLab}. As application of DCNNs becomes increasingly extensive, demand for model accuracy is also gradually increasing. Correspondingly, network architectures becomes more and more complex and deeper, leading to a sharp increase in model parameters and computational requirements. While deeper networks can improve accuracy, they also introduce efficiency issues, characterized by a trade-off between accuracy and efficiency. Traditional deep neural networks are quite challenging to deploy on mobile devices due to their high demands for storage and computing resources. As shown in Table 1, we compare storage usage and number of parameters for several classical DCNNs.

To augment the image processing capabilities and efficiency of mobile devices within the constraints of limited storage and energy consumption, the development of lightweight deep neural networks is paramount. These networks are specifically tailored for resource-constrained environments, achieving a streamlined transformation of deep learning models. This endeavor involves addressing the limitations imposed by energy consumption, computational power, and memory on such hardware. It also necessitates a reduction in storage demands and computational overhead through the application of innovative architectural designs and model compression techniques, without compromising the accuracy of the models.

\begin{table*}[t]  % 尝试将表格放置在合适的位置
  \centering  % 使表格居中
  \caption{Comparison of \#parameters and storage usage of DCNN models}  % 表格标题
  \label{tab:Table 1}  % 表格标签，用于引用
  \begin{tabular}{ccc}  % 定义一个有三列的表格
    \toprule
    DCNN Model & \#params & Storage Usage \\  % 表头
    \midrule
    AlexNet (2012) \cite{Krizhevsky2012AlexNet} & 60 million & 240 MB \\  % 第一行数据
    VGG16 (2014) \cite{Simonyan15VGG} & 138 million & 528MB \\  % 第二行数据
    VGG19 (2014) & 144 million & 575MB \\  % 第三行数据
    GoogLeNet (2014) \cite{Szegedy2015GoingDeeperWithConv-GoogLeNet} & 6 million & 24MB \\
    ResNet-50 (2015) \cite{He2016ResNet} & 25.5 million & 98MB \\
    ResNet-101 (2015) & 44.5 million & 178MB \\
    Inception-v4 (2016) \cite{Szegedy2017Inception-v4} & 42 million	& 160 MB \\
    DenseNet-201 (2017) \cite{Huang2017DenseNet}	& 20 million & 80 MB \\
    NASNet-A (2018) \cite{Zoph2018NASNet} & 88.7 million & 340 MB \\
    EfficientNet-B7 (2019) \cite{Tan2019EfficientNet} & 66 million & 250 MB \\
    \bottomrule
  \end{tabular}
\end{table*}

Extensive research has been conducted by both academia and industry on the lightweighting of DCNNs, resulting in notable progress. For example, Howard et al. introduced depthwise separable convolutions as a replacement for traditional convolutions, which facilitated the development of the MobileNet series, evolving from MobileNetV1 \cite{Howard2017MobileNetv1} through MobileNetV2 \cite{Sandler2018MobileNetv2} to MobileNetV3 \cite{Howard2019MobileNetv3}, with each iteration enhancing its design. In pioneering work, LeCun et al. \cite{LeCun1989OptimalBrain} proposed the "Optimal Brain Damage" strategy, which optimizes neural networks by pruning redundant connections, thereby accelerating training and enhancing generalization. Tan et al. \cite{Tan2019EfficientNet} advanced beyond manual design by employing reinforcement learning to develop Mobile Neural Architecture Search (MNAS), an automated approach to neural architecture optimization that efficiently generates CNN models tailored for mobile devices.

Current research on lightweight DCNNs primarily focuses on two aspects: rational architectural design and model compression techniques. Architectural design can be further divided into manual methods and AutoML-driven Neural Architecture Search (NAS). Model compression techniques mainly encompass pruning, knowledge distillation, weight quantization, and low-rank decomposition. These methods aim to reduce model size and enhance computational efficiency, catering to diversified application scenarios.

This paper is structured as follows: Section 2 will detail philosophy of rational architectural design in lightweight DCNNs, exploring strategies for optimizing structures and reducing parameters to build efficient and lightweight models. Section 3 will review assistance and impact of neural architecture search driven by AutoML on designing of lightweight DCNNs. Section 4 starts with several essential model compression techniques for deep neural networks, focusing on their application in lightweighting of DCNNs. Finally, we discuss current limitations in this field of research and propose prospects for future directions, aiming to promote further advancements in DCNN lightweighting.

\section{Improve Lightweight DCNNs Manually}

\subsection{Adjust Kernel Size}

In the field of convolutional neural networks, common convolution kernel sizes include 11×11, 7×7, 5×5, 3×3, and 1×1. Optimization of convolution kernel size primarily follows two directions: employing smaller kernels to replace larger ones, or using special "1×1 kernel" for network fine-tuning. LeCun et al.  \cite{Lecun1998Gradient} advocated in 1998 for the use of small convolution kernels (e.g., 3×3 kernel) as alternatives to larger kernels, believing that smaller kernels can reduce parameters while maintaining performance. Stacking multiple layers with small convolution kernels can gradually expand receptive field, achieving effects similar to larger kernels. Simonyan et al. \cite{Simonyan15VGG} highlighted benefits of multiple 3×3 kernels over larger kernels, which reduce parameters and increase network depth, thereby enhancing model performance. This approach leverages stacked layers with small kernels to capture fine-grained features and uses deeper networks to grasp complex patterns. He et al.  \cite{He2016ResNet} further promoted application of smaller kernels in ResNet, effectively training deeper networks with 1×1 and 3×3 kernels through residual connections and batch normalization. This method sustains model complexity while boosting performance and generalization capability. Szegedy et al.  \cite{Szegedy2016RethinkInception} proposed substituting a single 5×5 convolution layer with two consecutive 3×3 layers, reducing 28\% of parameters while maintaining receptive field, and analogously, replacing a 7×7 kernel with three sequential 3×3 layers, cutting approximately 45\% of parameters. Replacing single-layer large-scale convolutions with multi-layer small-scale ones enhances fitting capacity of DCNNs, augments discriminative power of decision functions, and realizes implicit regularization. Taking VGG as example, its adoption of small kernel strategy significantly reduced computational demands and parameter quantities, showcasing superior performance compared to other deep neural networks, demonstrating that this design concept not only optimizes model complexity but also excels in practical applications. Peng et al.  \cite{Peng2017LK}, on the other hand, suggested using two layers of 1×k and k×1 convolutions in place of k×k convolutions, significantly lowering computational requirements while maintaining a large receptive field.

In addition to using 3×3 kernels for parameter reduction, 1×1 kernels can also achieve the same goal. Szegedy et al. \cite{Szegedy2015GoingDeeperWithConv-GoogLeNet} introduced 1×1 kernels in Inception V3, effectively enhancing network performance and efficiency while significantly reducing parameters and computational burden. Furthermore, SqueezeNet, developed by Iandola et al. \cite{Iandola2016SqueezeNet}, extensively employs 1×1 kernels in place of 3×3 types, demonstrating that it achieves comparable accuracy with far fewer parameters than AlexNet. Given that 1×1 kernels operate on single pixels, they drastically prune network parameters, simplifying model complexity and accelerating training and inference. This dimension adjustment functionality enables adjustment of channel numbers without altering spatial dimensions of feature maps, facilitating adjustments to feature dimensions, which is crucial for modifying network width and decreasing parameter counts. GoogLeNet and ResNet leverage 1×1 convolutions to reduce channel numbers, thereby optimizing their models and alleviating computational loads. Within multi-channel feature maps, 1×1 convolutions act as cross-channel linear mixers, integrating inter-channel information to capture richer features. Depthwise separable convolutions in MobileNet exemplify this role, where 1×1 convolutions consolidate channel features to produce new representations, reducing computation by approximately 12\% compared to standard 3×3 convolutions, thus enhancing efficiency.

Small convolution kernels exhibit a series of pros and cons in deep convolutional neural network design. On the upside, smaller kernels curtail the number of parameters, decrease computational complexity and memory usage, and their stacking across multiple layers emulates large kernel receptive fields. This refinement enhances model's capacity to discern details, augmenting potential for nonlinear expression and deep feature learning. However, it is noteworthy that smaller kernel approaches often necessitate deeper network architectures to attain performance of larger kernels, which can escalate training difficulties. Moreover, excessive layers can amplify the risk of gradient vanishing or explosion, although regularization and proper initialization can alleviate these issues to some extent.

% \begin{figure}[htbp]  \centering  \includegraphics[height=4.5cm, width=7.5cm]{pdf_file}  \caption{Design}  \end{figure}

\begin{figure*}[t]
	\centering % 表示居中
	\includegraphics[width=\linewidth]{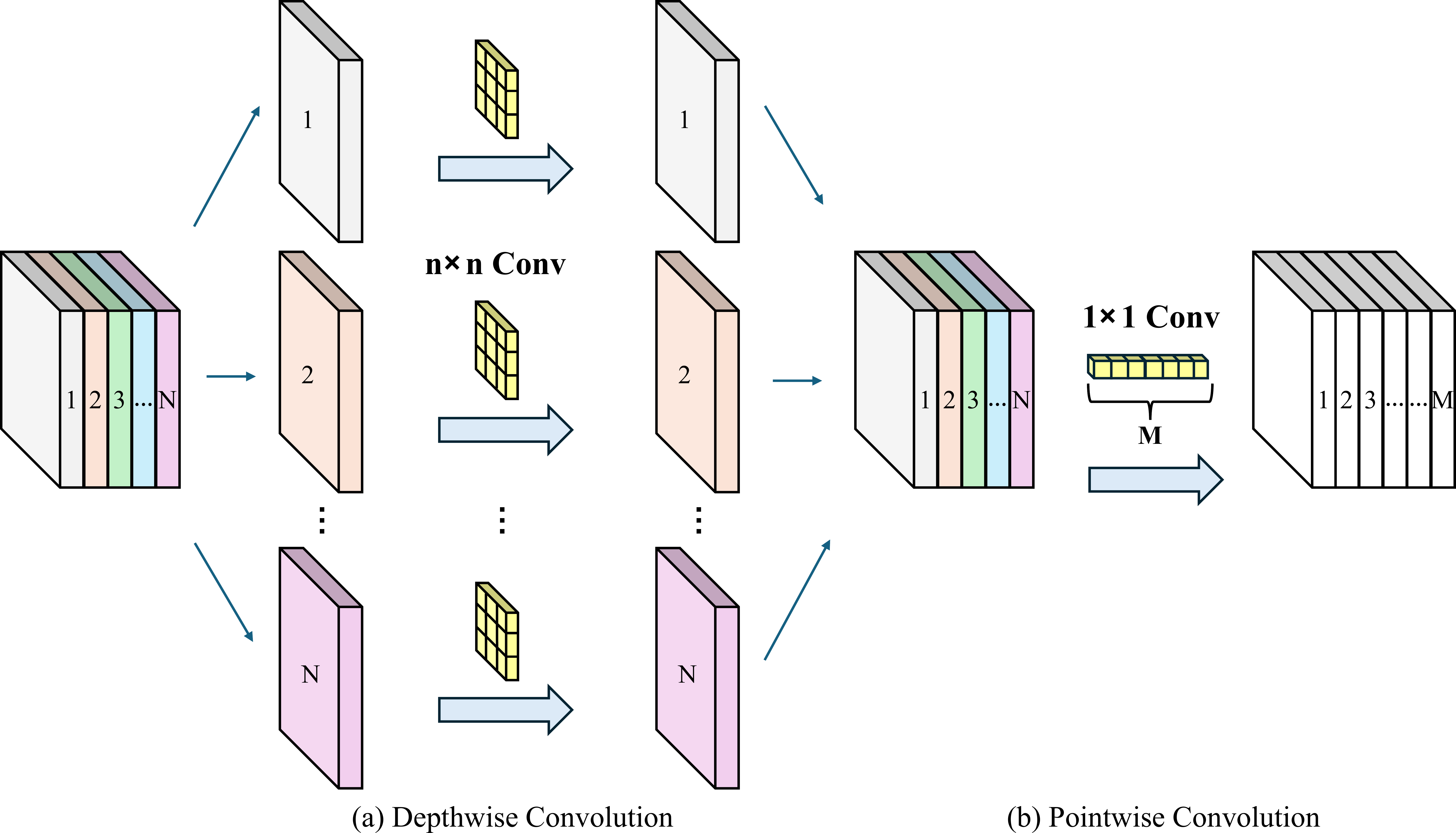}
	\caption{Depthwise Convolution serves on spatial feature extractor, operating on each channel independently. Meanwhile, Pointwise Convolution acts as a channel mixer, integrating information across various channels.}
    \label{tab: Figure 1}
\end{figure*}

\subsection{Improve Convolutional Structure}

Traditional convolution operations involve applying kernels to each input feature channel, accumulating through matrix multiplications to extract features, which frequently leads to model parameter expansion and redundancy in feature extraction phase. Consequently, researchers actively explore innovations in convolutional architectures aimed at reducing parameters meanwhile maintain accuracy. Depthwise separable convolutions and group convolutions, as two pivotal strategies, have significantly influenced and been widely implemented in this domain.

Depthwise separable convolutions  \cite{Howard2017MobileNetv1} ingeniously integrate depthwise convolution with pointwise convolution techniques. As shown in Figure 1.a, initially, depthwise convolution operates independently on each input feature channel, precisely extracting spatial features from feature maps. Subsequently, pointwise convolution employs 1×1 kernels to integrate information across channels, reinforcing feature interactions, which is shown in Figure 1.b. This two-stage strategy ensures efficient feature extraction while dramatically reducing parameters, facilitating model lightweighting.

The computational cost of depthwise separable convolutions is only $1/{k^2}$ of that of regular convolutions (where $k$ represents kernel size), significantly reducing computational complexity. Consequently, while maintaining accuracy, model size can be greatly reduced, which is highly valuable for practical deployment on resource-constrained devices.

Group convolution \cite{Krizhevsky2012AlexNet} was first introduced in AlexNet. Due to limitations in hardware, AlexNet innovatively split feature maps across multiple GPUs for parallel processing, merging results eventually to enhance resource utilization efficiency. Group convolutions divide input into several groups along the channels for independent convolution processes, subsequently recombining them to output features, thereby achieving a lightweight effect – if input features are divided into $g$ groups, parameter quantity is merely $1/g$ of the original amount. Nonetheless, group convolutions have limitations in that they restrict information exchange between channels, leading to output features not fully encompassing all input information. To address this shortcoming, it is common to incorporate pointwise convolutions from depthwise separable convolutions or channel shuffle strategies from ShuffleNet to facilitate information fusion. Compared to standard convolutions, group convolutions can reduce computational load to $1/g$, further minimizing memory usage and computational demands while maintaining high accuracy.

\begin{figure}[t]
	\centering % 表示居中
	\includegraphics[scale=0.1]{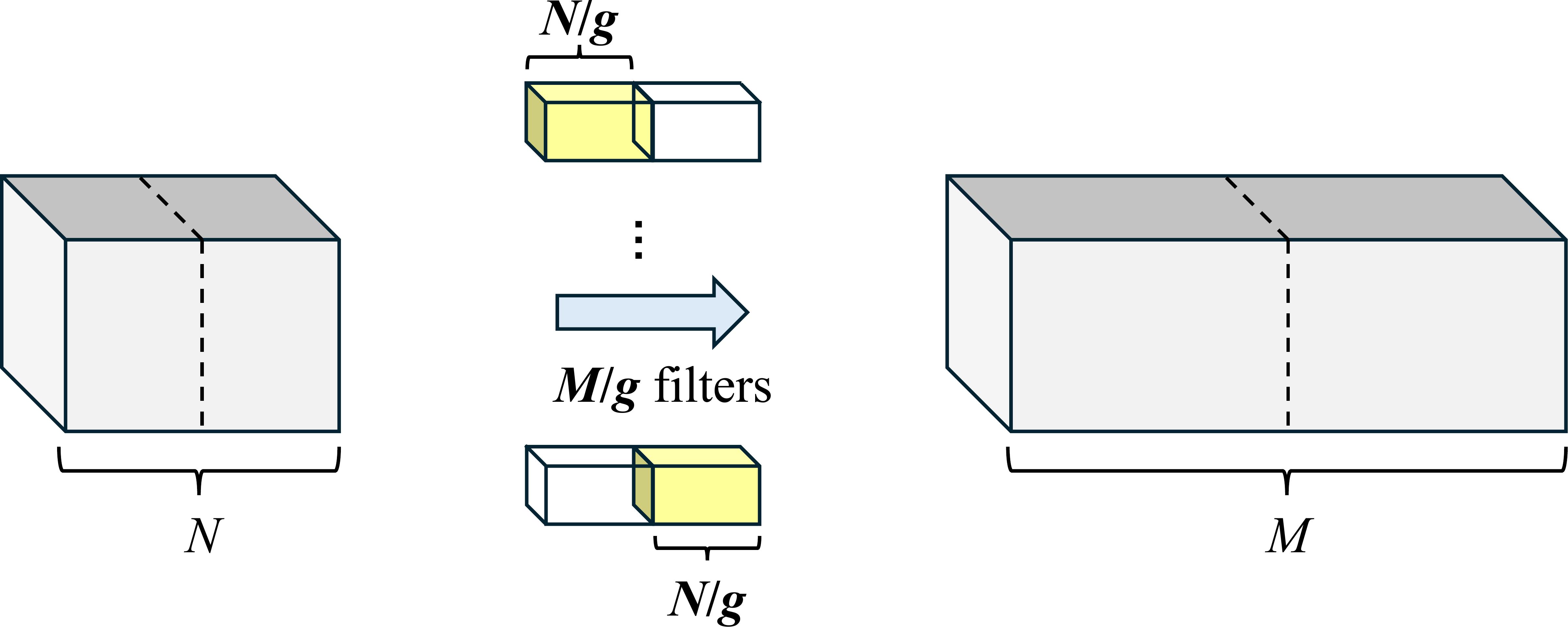}
	\caption{Group Convolution employs a set of kernels, each operating on its respective segment of the input.}
    \label{tab: Figure 2}
\end{figure}

Optimizing convolutional structures also encounters several limitations. On one hand, while improved methods such as depthwise separable convolutions and group convolutions effectively reduce parameter counts and computational burdens, they may undermine feature extraction capabilities, particularly exhibiting pronounced deficiencies when dealing with complex visual scenarios. On the other hand, these improvements rely on specific hardware acceleration, with limited optimization effects on general-purpose processors. Furthermore, model over-simplification can compromise robustness against adversarial noise and variations in data distributions to a certain degree, thereby threatening stability and reliability in applications.

\subsection{Use Cheaper Operations}

Multiplication operations exhibit significantly greater computational complexity than addition operations. In deep neural networks, the convolution operation, which is essentially a precise cross-correlation, assesses the similarity between input features and filters through extensive floating-point multiplications. To mitigate this, Chen et al. \cite{Chen2020AdderNet} introduced AdderNet, a model that substitutes traditional multiplications in convolutional neural networks with additions, thereby diminishing computational expenses and enhancing energy efficiency. AdderNet computes the L1 norm distance between input features and filters, circumventing the need for extensive floating-point multiplications. To refine the AdderNet training process, they engineered a specialized backpropagation algorithm and an adaptive learning rate strategy that modulates according to the gradient magnitude of each neuron. Empirical results indicate that AdderNet can achieve comparable accuracy to conventional CNNs on the ImageNet dataset, despite eschewing multiplications in its convolutional layers. Furthermore, Song et al. \cite{Song2021AdderSR} investigated the feasibility of AdderNet for single-image super-resolution (SISR) tasks. They identified challenges inherent in applying AdderNet to SISR and proposed mitigations, such as incorporating self-shortcut connections to bolster model efficacy and devising a learnable activation function to modulate feature distributions and refine detail rendition. These enhancements facilitate a substantial reduction in energy consumption for AdderNet-based SISR models, without compromising performance or visual quality relative to CNN benchmarks.

Xu et al. \cite{Xu2020PKKD} introduced Progressive Kernel Knowledge Distillation (PKKD), a technique designed to enhance the performance of AdderNet without incurring additional training parameters. This method commences by initializing a convolutional neural network identical in architecture to the teacher network. It then projects the features and weights of both the student network (AdderNet) and the teacher network into a novel space, thereby mitigating the challenge of accuracy deterioration. Conducting similarity assessments within a higher-dimensional space enables a precise differentiation between the two networks. Consequently, AdderNet can be systematically trained, leveraging information from both actual labels and the teacher network's predictions. Empirical evaluations reveal that AdderNet, when refined through PKKD, surpasses the original ResNet in terms of accuracy.

Han et al. \cite{Han2020GhostNet} introduced the Ghost module, a pioneering neural network component, constituting the GhostNet architecture. This architecture capitalizes on the observation that well-trained deep neural networks typically exhibit an abundance of feature maps, some of which are redundant, ensuring a comprehensive analysis of input data. The researchers identified that certain feature maps post the initial residual block of ResNet-50 exhibit pairwise similarities, implying that one map can be inexpensively derived from another, with the latter being termed a "ghost" of the former. Leveraging this insight, the authors advocate for a selective approach to feature map extraction from convolutional operations, suggesting that "ghost" feature maps can be generated through more economical operations. Accordingly, the Ghost module bifurcates each convolutional layer into two segments: the initial segment employs standard convolution with a stringent regulation on the number of feature maps to mitigate computational overhead; the subsequent segment produces additional feature maps, not through standard convolution, but via straightforward linear transformations.

\subsection{Find Equivalent Architecture}

In densely connected neural networks, merging is permissible for adjacent fully connected layers lacking an intervening nonlinear layer, condensing them into a singular fully connected layer. Given two fully connected layers with weights delineated by matrices $A$ and $B$, and an input vector $x$, the composite output is formulated as $y = B(Ax)$. By amalgamating these into matrix $C = BA$, the output simplifies to $y = Cx$, where C represents the weight matrix of the integrated fully connected layer. Numerous convolutional neural networks incorporate multi-branch architectures to enrich feature extraction. Nonetheless, these architectures frequently encounter computational inefficiency. To mitigate this, researchers have proposed structural reparameterization, drawing parallels to the merging of fully connected layers, thereby enhancing computational efficiency.

Structural reparameterization is a methodology that initially constructs a configuration of structures for the training phase and subsequently reformats them into a distinct configuration for inference, based on parameter equivalence. In practical applications, the focus is typically on the efficiency and cost of inference, given the typically ample availability of training resources. Consequently, it is common to employ more complex structures during training to attain superior accuracy and to induce beneficial attributes such as sparsity. For inference, however, more streamlined, yet structurally equivalent, structures are utilized to preserve these attributes. This approach implies a correspondence between the training structures, which are associated with one set of parameters, and the inference structures, which correspond to a different set of parameters.

Ding et al. \cite{Ding2021RepVGG} proposed RepVGG, an architecture that uses a single-branch topology during inference and a multi-branch topology during training. The multi-branch topology during training is transformed into a single-branch topology during inference through structural reparameterization, decoupling multi-branch topology during training from single-branch topology during inference. During training process, RepVGG uses identity and 1x1 branches, similar to ResNet, but these branches can be removed through structural reparameterization. Specifically, as shown in Figure 3, the 1x1 branch during training is regarded as a degenerate 3x3 convolution, and the identity branch is regarded as a degenerate 1x1 convolution, which is further regarded as a degenerate 3x3 convolution. The parameters of 1x1 branch and identity branch during training are converted into the parameters of 3x3 convolution layer during inference. In this way, the multi-branch model during training can be transformed into a single-branch model during inference through simple algebraic transformations. After structural reparameterization, RepVGG uses only 3x3 convolutions and ReLU operations during inference, greatly simplifying inference process and improving inference speed.

\begin{figure}
	\centering % 表示居中
	\includegraphics[scale=0.15]{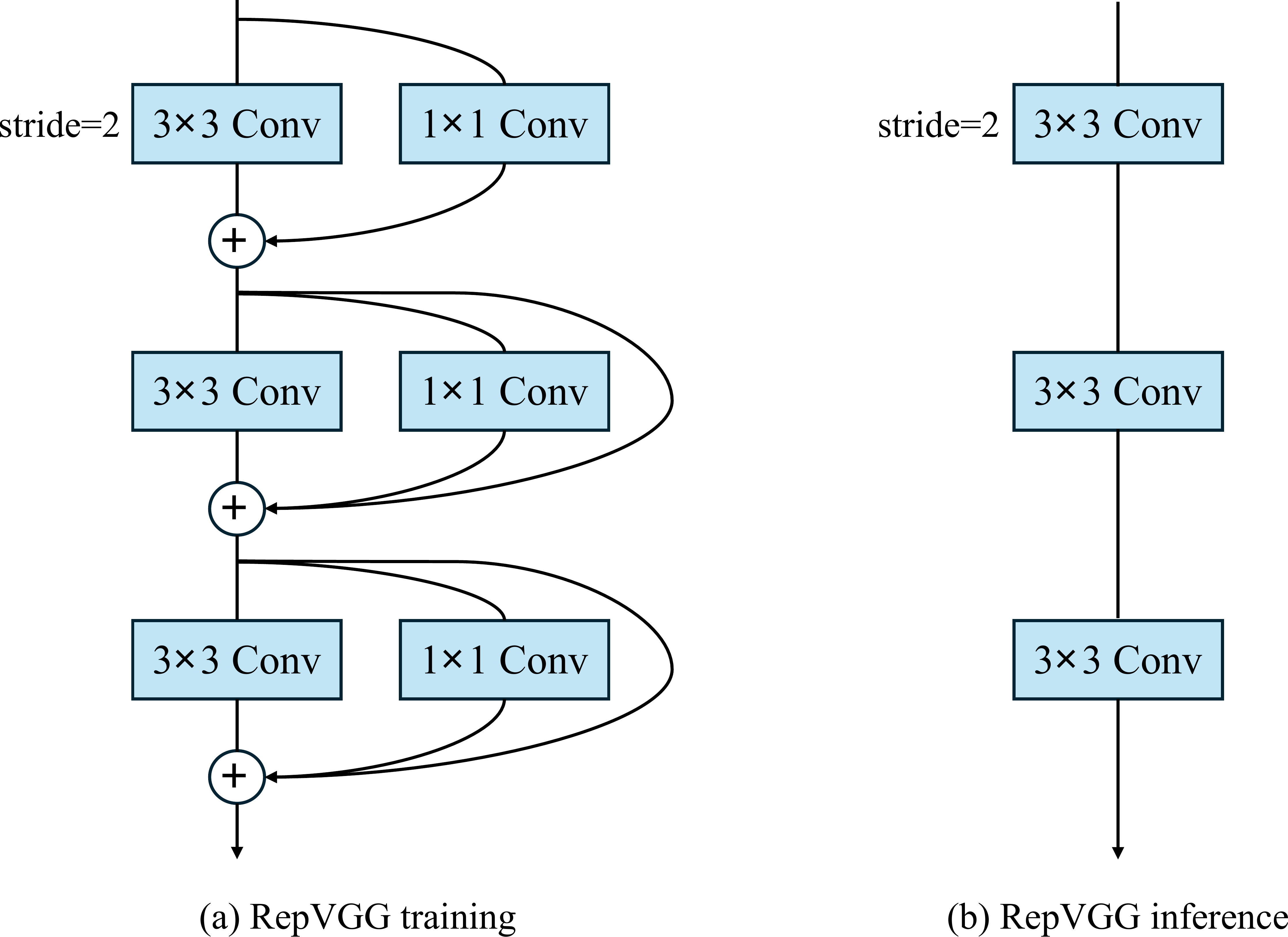}
	\caption{In RepVGG, a multi-branch structure with residual connection is employed during the training phase, whereas a chain structure equipped with 3×3 kernel is utilized for inference.}
    \label{tab: Figure 3}
\end{figure}

Structural reparameterization can not only accelerate inference but also enable feature reuse without concatenation operations. Chen et al. \cite{Chen2022RepGhost} proposed a hardware-efficient RepGhost module. They noted that although traditional concatenation operations are free in terms of parameter and FLOP counting, they incur high computational costs on hardware devices in deployment. Therefore, RepGhost module removes inefficient concatenation operations. Then, ReLU activation function is moved after depthwise convolution layer to comply with the rules of structural reparameterization. Subsequently, during feature fusion, different layers' features are combined using addition operations rather than concatenation in the feature space. In this way, feature reuse process can be transferred from feature space to weight space during inference, enabling implicit and efficient feature reuse. Finally, since all operations are linear functions, the final fusion process can be performed directly in weight space without introducing any additional inference time, making the architecture more efficient.

\section{Design Lightweight DCNNs Automatically Driven by AutoML}

The efficacy of deep learning is primarily due to its end-to-end learning methodology and its capacity for automated feature extraction. Nonetheless, the escalating volume of data and the relentless pursuit of enhanced performance have rendered the manual design of deep learning models increasingly arduous. This complexity demands considerable domain-specific knowledge and expertise from designers, while also entailing substantial temporal and laborious investments. Such demands severely restrict the deployment and evolution of lightweight networks on mobile platforms. AutoML, an emerging research discipline, endeavors to automate the machine learning workflow through algorithmic approaches, with Neural Architecture Search being a pivotal research avenue. Consequently, extensive investigation into neural architecture search is instrumental in discovering more efficient and swiftly learning network architectures. NAS, powered by AutoML, can substantially reduce the design burden and offer innovative insights and directives for the lightweight exploration of DCNNs. Empirical evidence has demonstrated NAS's superior performance in computer vision tasks, including image classification and semantic segmentation, frequently surpassing the capabilities of manually crafted neural network models. Essentially, neural architecture search can be conceptualized as an optimization challenge within a high-dimensional parameter space, which researchers dissect into three distinct sub-problems: the construction of the search space, the establishment of the search strategy, and the selection of the performance evaluation mechanism.

\begin{enumerate}
    \item Search space is the set of candidate structures available for selection in optimization problem of neural network architecture, that is, the domain of solutions. It encompasses all possible network architectures, from simple layer stacking to complex inter-layer connections and branching structures.
    \item Search strategy is the method or algorithm used to find optimal network architecture within the search space. It determines how to effectively traverse the search space to find network architectures that meet performance requirements.
    \item Performance evaluation strategies are used to assess the performance of each candidate network architecture in search space. They provide a standard or metric to measure the superiority or inferiority of different architectures.
\end{enumerate}

The schematic overview of NAS is depicted in Figure 4. The search strategy delineates a neural network model $N$ from a pre-defined search space $S$, subsequently assessing its performance via a designated evaluation protocol. This performance metric is then conveyed as a reward signal to the search strategy. Subsequently, the search strategy refines the neural network model, leveraging the accrued rewards, in an iterative pursuit of the optimal model configuration.

\begin{figure}
	\centering % 表示居中
	\includegraphics[scale=0.15]{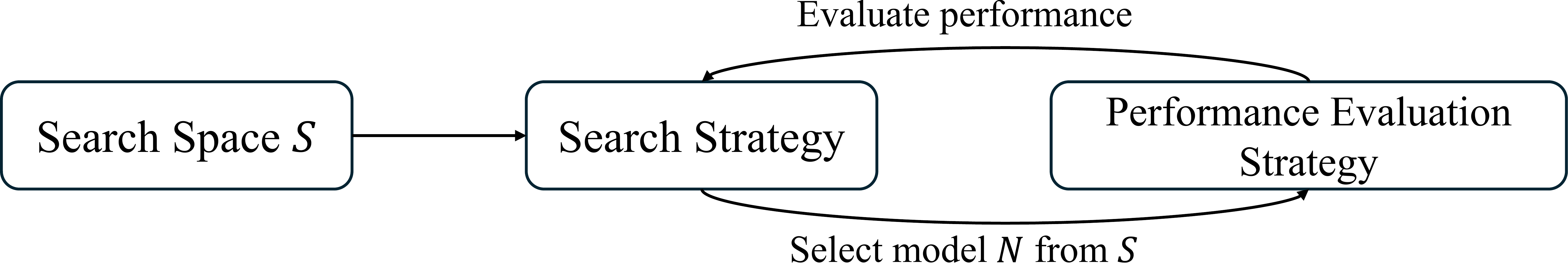}
	\caption{General Process of Neural Architecture Search.}
    \label{tab: Figure 4}
\end{figure}

\subsection{Search Space}

Search space is an indispensable core element in NAS, representing a subset of the comprehensively defined neural network models and offering the foundational elements for neural network construction. Selecting an appropriate search space is pivotal as it streamlines the optimization challenge; a judiciously crafted search space diminishes the search problem's complexity and substantially augments the search efficiency, increasing the probability of identifying the optimal solution within a constrained time-frame. Consequently, the choice of search space is critical to the success of the entire network search endeavor, with a superior search space often correlating with exceptional search outcomes. Search spaces can be categorized into global and local domains based on the distinctive attributes of network architectures. The global search space concentrates on the holistic network architecture, seeking to identify network configurations with globally optimal performance. Conversely, the local search space targets specific segments or components within the network, aiming to elevate overall performance through localized structural optimization. Each search space type possesses unique advantages and is tailored to distinct application contexts and requirements, affording researchers the flexibility to make informed selections based on specific scenarios.

\subsubsection{Global-based Search Space}

In the manual design of neural network layers, the parameters encompass the dimensions and quantity of convolutional kernels, convolution types, stride lengths, the count of connected layers, and the presence of residual connections. NAS systematically identifies these hyperparameters for each layer, effectively conducting a comprehensive exploration of the network's architecture. This exploration constitutes the global search space, which, contingent upon the network's configuration, can be differentiated into serial and intricate multi-branch structures.

Figure 5 shows the search space of chain and multi-branch structures, where different nodes represent different types of layers in the network, indicated by different colors. The directed edges in the figure represent the relationships between input nodes and output nodes. The chain structure is illustrated in Figure 5.a, where each hidden layer of the neural network is only connected to the two adjacent layers, with no cross-layer connections. Some classic networks, such as VGG and LeNet, adopt a chain structure.

\begin{figure*}
	\centering % 表示居中
	\includegraphics[width=\linewidth]{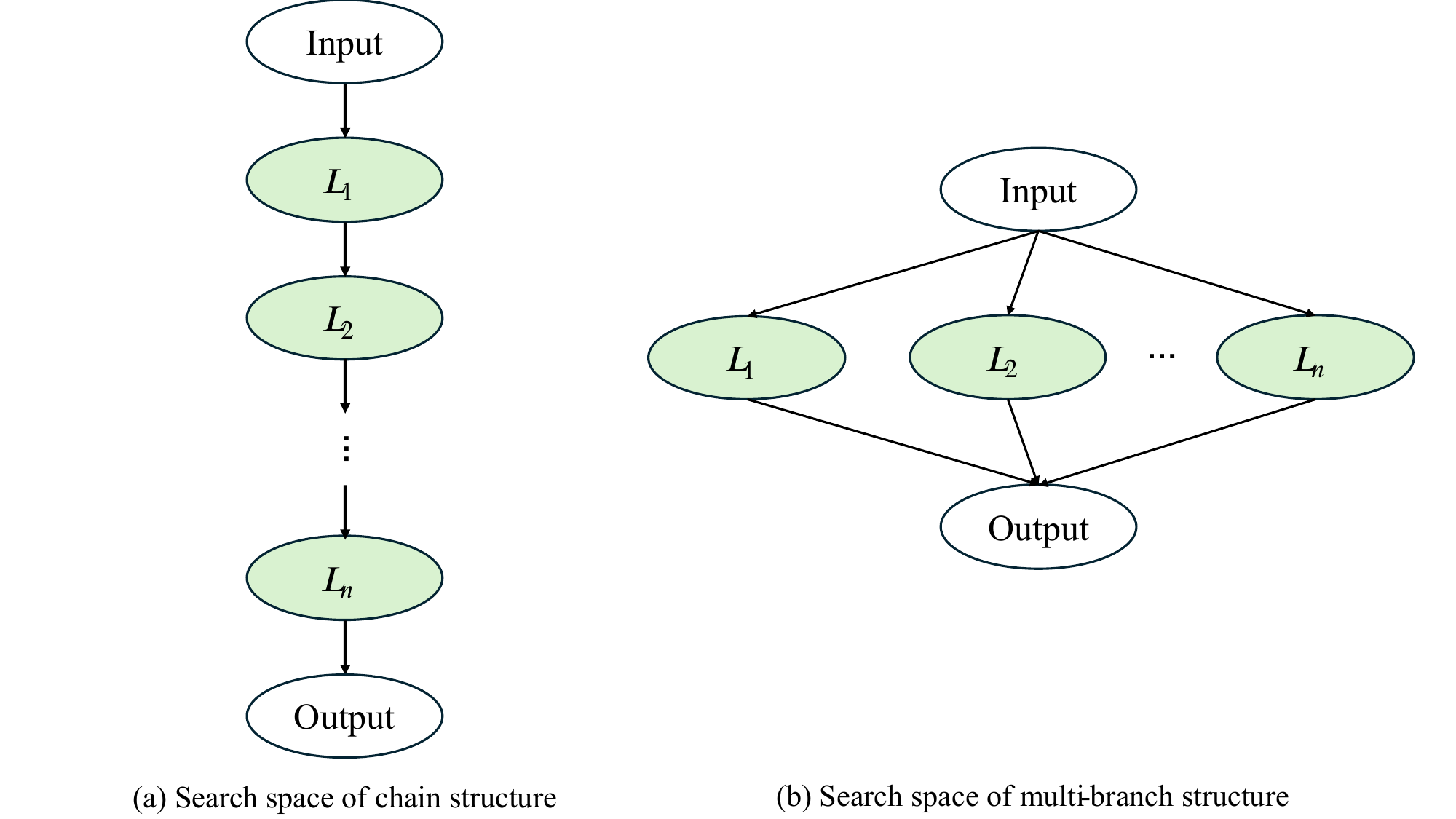}
	\caption{Brief showcase of chain structure and multi-branch structure.}
    \label{tab: Figure 5}
\end{figure*}

Baker et al. \cite{Baker2016NASRL} first proposed the concept of chain-structured search space, setting operations such as convolution, pooling, and fully connected layers, while also considering certain special constraints to exclude a portion of significantly incorrect neural network architectures. They also imposed limits on the number of network layers and operations to prevent the network structure from becoming overly complex. Unlike Baker et al., Zoph et al. \cite{Zoph2018NASNet} added residual connections to the chain structure, increasing the flexibility of the search space and expanding the size of the chain-structured search space.

The simplicity of chain structures often precludes their ability to effectively encapsulate and model the intricate, multi-level, and non-linear interrelationships inherent in complex data, thereby constraining the model's expressiveness and precision. Furthermore, chain structures are prone to gradient vanishing or explosion, which impedes the training of deep networks. In contrast, multi-branch structures segment input features across various branches, facilitating a more efficient capture of multi-scale features and making them well-suited for tackling multi-scale variations and occlusions in pedestrian detection. The dual pooling attention mechanism augments the model's capacity to concentrate on salient features while mitigating the impact of irrelevant ones, thus enhancing detection precision. Empirical evidence suggests that multi-branch network architectures substantially outperform traditional single-branch networks in detection tasks, particularly in complex environments. The advent of multi-branch structures effectively rectifies the limitations of chain structures by incorporating branching and cross-layer connectivity, fostering feature diversification, bolstering the model's generalization and computational capabilities, and ameliorating the challenges of gradient-related phenomena, positioning it as a prevalent network architecture.

\subsubsection{Local-based Search Space}

With in-depth exploration of multidimensional optimization problem of network search technology, researchers have realized that the approach of directly modeling and searching the entire network in the global search space faces significant computational cost challenges. This is primarily due to the complexity of modern neural networks, which often contain hundreds of convolutional layers, each with various types and hyperparameter choices. Particularly when dealing with large-scale datasets, the search space becomes exceptionally vast, leading to a sharp increase in computational costs. To address the aforementioned issues, researchers have been inspired by the presence of numerous repetitive Block structures in manually designed deep neural networks. By stacking these Block structures to form neural network architectures, this design not only ensures excellent performance of the neural network architecture but also allows for the convenient application of the searched neural network to different datasets and tasks by flexibly adjusting the parameters, demonstrating outstanding generalization capabilities.

Zoph et al. \cite{Zoph2018NASNet} were the first to explore block-based search spaces and proposed the search space for MNASNet. As shown in Figure 6, different blocks can have different internal layer structures, while the layer structures within the same block are essentially the same. During the search process, it is only necessary to determine the operations and connections for each block. Based on MNASNet, researchers further explored block-based search spaces. Liu et al. \cite{Liu2018HRENAS} proposed a hierarchical representation method for neural networks. This method first searches within a basic set of operations to obtain various different units. Then, these initially searched units are used as basic building blocks for further searching to construct the final hierarchical network structure. This method achieves a hierarchical representation of neural network structures through a step-by-step construction and search approach, which helps improve network performance and optimize search efficiency. Real et al. \cite{Real2020AutoMLZero} refined the search granularity from the perspective of search units, selecting smaller basic mathematical operations as search units, such as matrix multiplication and tensor scaling. This finer granularity in the selection of search units aids in achieving more precise control and optimization during the neural network search process, thereby enhancing search efficiency and improving network performance.

\begin{figure*}
	\centering % 表示居中
	\includegraphics[width=\linewidth]{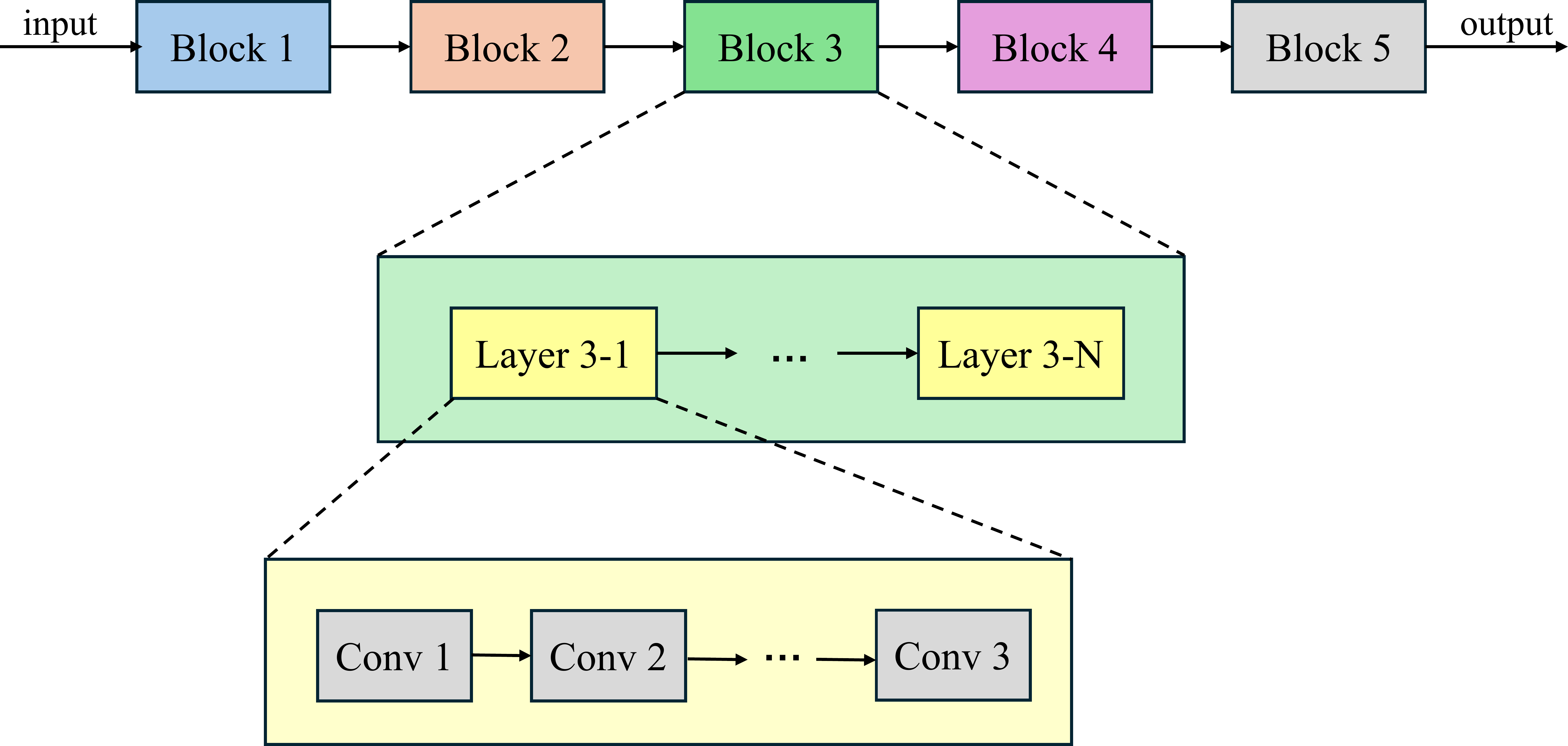}
	\caption{In MNASNet, different block may have diverse internal structures, resulting in a local-based search space.}
    \label{tab: Figure 6}
\end{figure*}

In addressing optimization challenges, the deployment of global and local search spaces offers distinct benefits and constraints. Global search algorithms excel in traversing the entire search space, thereby circumventing entrapment in local optima and potentially attaining the global optimum. Such algorithms are optimally suited for complex problems characterized by numerous local optima. Nevertheless, global searches are frequently resource-intensive, particularly when confronted with expansive or intricately structured search spaces. Conversely, local search algorithms are prized for their efficiency and swift convergence towards optimal solutions within confined domains. These are adept for refinement tasks in proximity to the optimal solution but are susceptible to entrapment in local optima, particularly within rugged or multi-modal search landscapes. 

In conclusion, to maximize the benefits of diverse search methodologies, the implementation of a hybrid strategy is typically essential. This strategy entails leveraging global search techniques to evade local optima, concurrent with the application of local search techniques to expedite the refinement of specific solutions. The judicious amalgamation of these methodologies ensures the comprehensiveness of solutions while concurrently enhancing search efficiency, thereby yielding superior optimization outcomes.

\subsection{Search Strategy}

After defining the search space, selecting an appropriate search strategy is crucial. An effective strategy not only guides the rapid construction of high-performing models but also helps prevent premature convergence to sub-optimal architectures. This section introduces two primary search strategies: reinforcement learning-based and gradient-based methods.

\subsubsection{Search Strategy Based on Reinforcement Learning}

Reinforcement learning is a branch of machine learning that describes how an agent learns a policy to maximize rewards or achieve specific goals through interaction with the environment. The process of reinforcement learning involves the agent performing actions in the environment, observing changes in the environment, and continuously improving its policy based on the rewards received to adapt to future tasks. Essentially, it aims to find a suitable policy function, as illustrated in Figure 7.

\begin{figure}[t]
	\centering % 表示居中
	\includegraphics[scale=0.15]{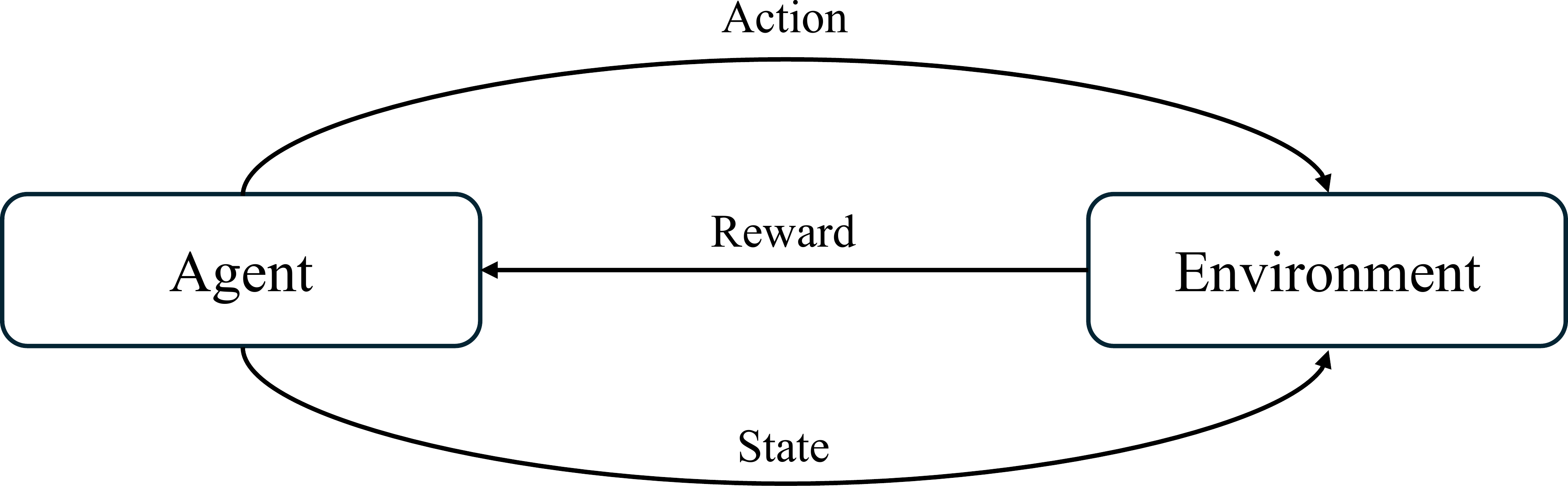}
	\caption{General process of Reinforcement Learning.}
    \label{tab: Figure 7}
\end{figure}

Bello et al. \cite{Bello2017NOSRL} proposed an innovative method for generating neural network architectures based on a recurrent network trained with reinforcement learning, aimed at significantly improving the accuracy of the generated architectures. This method has the capability to design models from scratch, allowing its performance on tasks such as image recognition and language modeling to rival or even surpass that of the best manually designed architectures. In this approach, the generated architectures are treated as variable-length token sequences, while the recurrent network acts as a controller responsible for generating these architectures. After training and evaluating the generated architectures, a reward signal is provided to the controller as feedback. The parameters of the controller are iteratively updated using reinforcement learning rules, thereby achieving continuous optimization.

Baker et al. \cite{Baker2016NASRL} proposed a method for designing neural network architectures based on the Q-learning algorithm, which transforms the learning task into a Markov decision process without the need for an environmental model, allowing the agent to complete tasks without estimating environmental dynamics. The agent selects layers using the $\epsilon$−greedy policy and adjusts the $\epsilon$ value over time. Experience replay techniques are used to accelerate convergence, optimizing Q-values through iterative update equations. Key parameters of the algorithm include the learning rate $\alpha$ and the discount factor $\gamma$ , which determine the weight of new information relative to old information, as well as the trade-off between short-term and future rewards. Under infinite sampling, the algorithm can converge to the optimal solution. The network generated by the agent achieves accuracy on CIFAR-10 comparable to or better than highly tuned manually designed networks, while also reducing the total number of parameters in certain cases.

Reinforcement learning-based search strategies are effective methods for solving complex decision-making problems, as they learn how to optimize decisions through the interaction between agents and the environment. The advantage of such strategies lies in their ability to adaptively learn optimal policies without the need for pre-defined detailed rules, making them particularly suitable for situations with dynamic environmental changes or incomplete prior knowledge. The key to reinforcement learning is the design of effective reward mechanisms and state representations, which directly affect the learning outcomes and the performance of the policies. However, reinforcement learning also faces challenges, such as the curse of dimensionality in state spaces, the sparsity and delay of rewards, and the high computational costs during the training process. Nevertheless, with the development of deep learning technologies and the enhancement of computational resources, reinforcement learning-based search strategies have demonstrated significant potential and broad application prospects in various fields.

\subsubsection{Gradient-based Search Strategy}

Reinforcement learning-based search strategies are not without limitations. The inapplicability of gradient information due to the non-differentiability of the process hinders the optimization of the search, leading to protracted training periods and diminished efficiency. In contrast, gradient-based search methodologies are designed to harness gradient information to create a differentiable search space, thereby enhancing the efficiency of the search process.

Liu et al. \cite{Liu2018DARTS} proposed a new algorithm called DARTS (Differentiable Architecture Search) for efficient architecture search. This algorithm addresses the scalability challenge by expressing tasks in a differentiable manner. Unlike traditional methods that rely on non-differentiable search techniques, DARTS utilizes continuous relaxation of architecture representation and employs gradient descent for efficient architecture search. Furthermore, the architectures learned by DARTS on the CIFAR-10 and PTB datasets have been shown to transfer to ImageNet and WikiText-2, respectively. The algorithm also achieves significant efficiency improvements, reducing the cost of architecture search to a few GPU days, attributed to the use of gradient-based optimization; however, DARTS suffers from excessive memory usage. To address this issue, Cai et al. \cite{Cai2019ProxylessNAS} proposed the ProxylessNAS method, which significantly reduces memory consumption, allowing for the search of large-scale candidate sets with the same computational resources as conventional training, and achieving excellent performance on the CIFAR-10 and ImageNet datasets. Wu et al. \cite{Wu2019FBNet} introduced a method based on differentiable neural architecture search called FBNet, aimed at designing hardware-aware efficient ConvNets. By optimizing the architecture distribution of ConvNets, FBNet can optimize ConvNet architectures without enumerating and training individual architectures. FBNet outperforms both manually designed and automatically generated state-of-the-art models in terms of performance, while achieving higher accuracy and lower latency.

The above neural architecture search methods mainly target GPUs or smartphones. However, memory resources of microcontrollers are two to three orders of magnitude smaller than those of smartphones. These methods do not consider the memory constraints of microcontrollers, leading to models that cannot run on these resource-constrained devices. Therefore, Lin et al. proposed MCUNet \cite{Lin2020MCUNet}, which is a system-algorithm co-designed framework for addressing deep learning on microcontrollers. MCUNet primarily consists of two parts: TinyNAS (for optimizing neural network architectures) and TinyEngine (for improving inference efficiency).

TinyNAS automatically optimizes the search space by analyzing the computational distribution that satisfies the model. Specifically, TinyNAS generates different search spaces by scaling the input resolution and model width. Each search space configuration contains 3.3 × $10^{10}$ possible subnetworks. To evaluate the quality of the search space, TinyNAS randomly samples $m$ networks and compares their FLOPs distribution CDF. Since accuracy is typically proportional to computation within the same model family, TinyNAS selects search spaces that are more likely to produce high FLOPs models under memory constraints.

TinyEngine optimizes memory scheduling based on the entire network topology to increase input data reuse, reduce memory fragmentation, and minimize data movement. Additionally, an in-place depthwise convolution method is proposed, further reducing peak memory usage. Depthwise convolutions do not require cross-channel filtering, so once the computation for one channel is complete, the input activations for that channel can be overwritten and used to store the output activations of another channel.

Subsequently, Lin et al. proposed MCUNetV2 \cite{Lin2021MCUNetV2}. They identified the memory bottleneck as the limiting factor for deploying deep learning on microcontroller units, and MCUNetV2 aims to address the imbalance in memory distribution present in existing deep learning systems. MCUNetV2 tackles this issue through the introduction of patch inference scheduling and receptive field redistribution. Specifically, patch inference scheduling allows the model to operate on smaller spatial regions of the feature map, thereby significantly reducing peak memory usage. However, this approach introduces problems related to overlapping patches and computational overhead. To mitigate this overhead, MCUNetV2 further proposes a receptive field redistribution method, which shifts the receptive fields and workload to later stages of the network by reducing the initial receptive fields and increasing them at later stages. This enables the maintenance of model accuracy while reducing computational overhead.

\begin{table*}[t]  % 尝试将表格放置在合适的位置
  \centering  % 使表格居中
  \caption{Comparison of different performance evaluation strategies}  % 表格标题
  \label{tab:Table 2}  % 表格标签，用于引用
  \begin{tabular}{ccc}  % 定义一个有三列的表格
    \toprule
    Performance Evaluation Strategy & Core Idea & References \\  % 表头
    \midrule
    Low fidelity & \begin{tabular}{p{10cm}}
    By adopting simplified evaluation methods or reducing evaluation accuracy, poorly performing candidate network structures can be quickly screened out, thereby avoiding excessive waste of computational resources on them. This approach can significantly reduce the computational load during the search process while maintaining search effectiveness, thus accelerating the search process.
    \end{tabular} & [44-46] \\  % 第一行数据
    Surrogate model & \begin{tabular}{p{10cm}}
    Train the agent model using a simplified approximate task based on actual tasks, evaluate the performance of the agent model, and finally transfer the best agent model to the actual task.
    \end{tabular} & [47-49] \\  % 第二行数据
    Network Morphism & \begin{tabular}{p{10cm}}
    Efficiently transforming or adjusting the network structure while maintaining the network functionality. This transformation is typically aimed at optimizing network performance, simplifying the network structure, adapting to different datasets or tasks, while ensuring the consistency of the network's output or behavior.
    \end{tabular} & [50-52] \\  % 第三行数据
    One-time Search & \begin{tabular}{p{10cm}}
    By automating the design of neural network structures, the optimal neural network architecture can be found in one go, thereby reducing the workload of manual design.
    \end{tabular} & [53-55] \\ % 第四行数据
    \bottomrule
  \end{tabular}
\end{table*}

Gradient-based search strategies, such as gradient descent and its variants (e.g., stochastic gradient descent, Adam, etc.), are widely used in the fields of machine learning and deep learning for optimization problems. These strategies iteratively update parameters by calculating the gradient of the objective function with respect to the parameters, aiming to minimize or maximize the objective function. Advantages include simplicity of implementation and high computational efficiency, making them particularly suitable for handling large-scale datasets. However, they also face limitations such as the tendency to get trapped in local minima, sensitivity to parameter initialization and learning rate selection, and potential difficulties in achieving global optimality when dealing with non-convex problems. Despite these challenges, by introducing advanced optimization techniques and adjusting strategies, gradient-based methods continue to demonstrate their strong performance and wide applicability in practical applications.

\subsection{Performance Evaluation Strategy}

Performance evaluation strategies are mainly used to assess the quality of the network structure after it has been searched, guiding adjustments to the search strategy. The evaluation process is very time-consuming and requires a large amount of GPU computing resources. Therefore, designing efficient and reasonable performance evaluation strategies is particularly important.Researchers have conducted studies on how to efficiently and accurately measure the performance of neural network models, proposing methods such as low-fidelity evaluation, surrogate model evaluation, network morphism, and one-shot search. The comparison of different performance evaluation methods is shown in Table 2.

\section{Model Compression for DCNNs}

The process of compressing deep learning models entails the elimination of superfluous parameters within neural networks and the streamlining of their architecture. This optimization results in a more concise and efficient model variant that maintains equivalent performance. Consequently, these refined models exhibit reduced computational and memory requirements, enhancing their versatility compared to their unoptimized counterparts and rendering them apt for deployment in edge devices.

In addition to manually designing lightweight networks, researchers are also exploring other ways to achieve lightweight models, among which model compression is widely favored. He et al. \cite{He2018AMC} introduced model acceleration based on AutoML, achieving efficient model compression results in an automated manner. Shuvo et al. \cite{Shuvo2023EAReview} focused on four research directions for efficient DL inference on edge devices: (1) novel DL architecture and algorithm design; (2) optimization of existing DL methods; (3) development of algorithm-hardware co-design; (4) design of efficient accelerators for DL deployment. Han et al. \cite{Han2016DeepCompression} proposed a deep compression method (i.e., first pruning the network by only learning important connections, then quantizing weights to enforce weight sharing, and finally applying Huffman coding) to reduce the storage requirements of neural networks by 35 to 49 times without affecting accuracy.

According to the redundancy of neural networks in different aspects, different model compression algorithms could be performed, which can be roughly divided into model pruning, knowledge distillation, weight quantization, and low-rank decomposition. A comparison of these methods is conducted based on the different characteristics of model compression methods, as shown in Table 3.

\begin{table*}[htbp]  % 尝试将表格放置在合适的位置
  \centering  % 使表格居中
  \caption{Comparison of different model compression methods}  % 表格标题
  \label{tab:Table 3}  % 表格标签，用于引用
  \begin{tabular}{cccc}  % 定义一个有三列的表格
    \toprule
    Compression Methods & Core Mechanism & Advantages & Limitations \\
    \midrule
    Model Pruning & \begin{tabular}{p{3.7cm}}
    Remove non-critical connections in neural networks to reduce the number of parameters and the complexity of the model.
    \end{tabular} & \begin{tabular}{p{3.7cm}}
    Reduce computational burden and storage requirements, and improve inference speed.
    \end{tabular} & \begin{tabular}{p{3.7cm}}
    May lead to a decline in performance, and the degree of pruning is difficult to balance.
    \end{tabular} \\ % 第一行
    Weight Quantization & \begin{tabular}{p{3.7cm}}
    Reduce the representation precision of model weights by converting floating-point numbers to low-bit-width fixed-point numbers or integers.
    \end{tabular} & \begin{tabular}{p{3.7cm}}
    Significantly reduce model size and improve operational efficiency.
    \end{tabular} & \begin{tabular}{p{3.7cm}}
    Precision loss, especially in low-bit-width quantization.
    \end{tabular} \\ % 第二行
    Low-rank Approximation & \begin{tabular}{p{3.7cm}}
    Reducing the number of parameters by decomposing the weight matrix into a low-rank format.
    \end{tabular} & \begin{tabular}{p{3.7cm}}
    Reducing parameters and computational costs.
    \end{tabular} & \begin{tabular}{p{3.7cm}}
    May lead to a decrease in model capacity and accuracy.
    \end{tabular} \\ %
    Knowledge Distillation & \begin{tabular}{p{3.7cm}}
    Training a small network (student) to mimic the behavior of a large network (teacher).
    \end{tabular} & \begin{tabular}{p{3.7cm}}
    Reduce the model size while maintaining performance effectively.
    \end{tabular} & \begin{tabular}{p{3.7cm}}
    The distillation process can be complex and time-consuming.
    \end{tabular} \\ %
    \bottomrule
  \end{tabular}
\end{table*}

\subsection{Model Pruning}

Model pruning, as a common method of model compression, is widely used in convolutional neural networks. Given the redundancy of parameters in neural networks, by reasonably pruning a certain proportion of parameters, the model size can be effectively reduced without compromising network performance. The process of model pruning is shown in Figure 8.

\begin{figure}[htbp]
	\centering % 表示居中
	\includegraphics[scale=0.135]{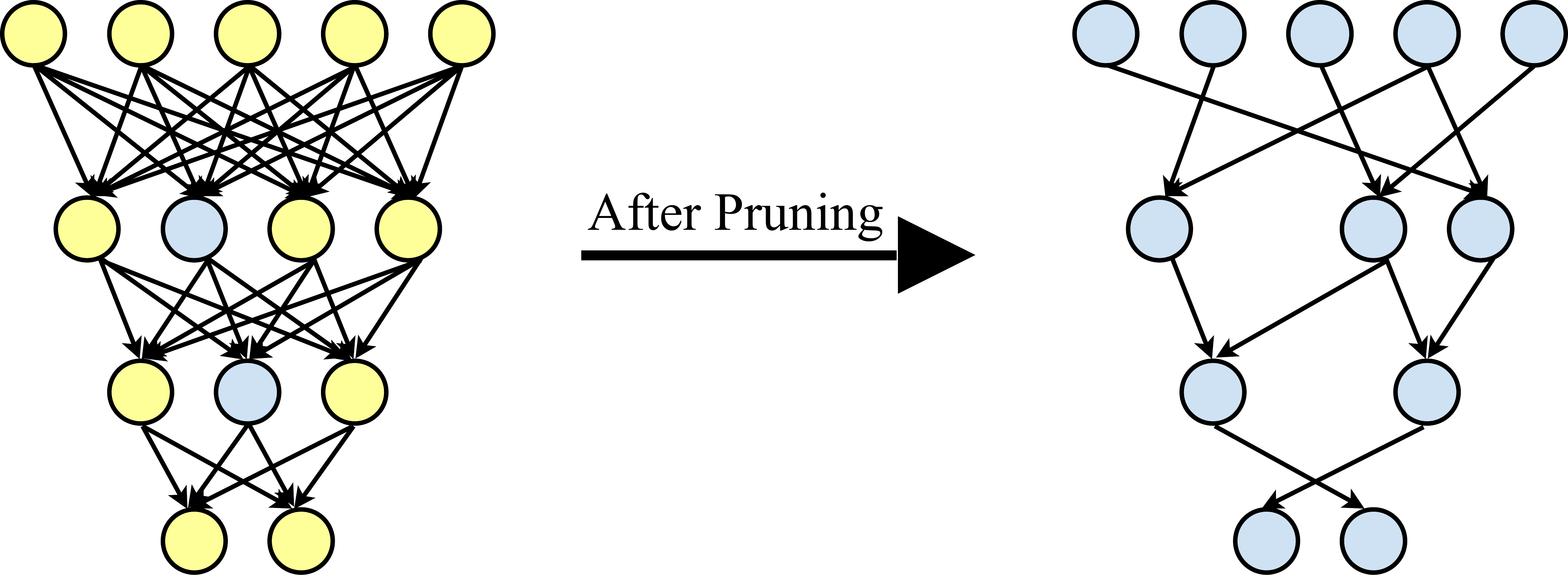}
	\caption{Brief showcase of Model Pruning.}
    \label{tab: Figure 8}
\end{figure}

Model pruning primarily divides into structured pruning and unstructured pruning. Unstructured pruning removes insignificant neurons, and correspondingly, the connections between pruned neurons and other neurons are ignored during computation. Since the model after unstructured pruning is typically sparse and disrupts the original structure of the model, this method is also known as fine-grained pruning. Unstructured pruning can significantly reduce the number of model parameters and theoretical computational load, but existing hardware architectures, especially edge devices with constrained resources, cannot get acceleration benefit from unstructured pruning due to sparse models. In contrast to unstructured pruning, structured pruning typically operates at the level of filters or entire network layers. When a filter is pruned, the previous feature map and the next feature map undergo corresponding changes, but the model structure remains intact, still allowing acceleration through GPUs or other general hardware.

Zhuang et al. \cite{Zhuang2018Discrimination-awareChannelPruning} proposed a novel discriminative perceptual channel pruning method. Unlike existing pruning methods that train from scratch and impose sparsity constraints on channels, this pruning method introduces additional loss into the network to enhance the discriminative power of intermediate layers. It then selects the most discriminative channels for each layer by considering additional reconstruction errors, achieving significant theoretical breakthroughs. Experimental results demonstrate that on ILSVRC-12, the pruned ResNet-50 channels were reduced by 30\%, and the top-1 accuracy was even 0.39\% higher than that of the original model.

Luo et al. \cite{Luo2017EntropyBasedCNNCompress} proposed an entropy-based CNN compression pruning method to assess the importance of parameters in convolutional kernels, achieving significant performance improvements by accelerating and compressing existing CNN models through a convolutional kernel pruning strategy. Yang et al. \cite{Yang2017Energy-AwarePruning} introduced an energy-aware pruning algorithm for CNNs that directly utilizes the energy consumption of the CNN to guide the pruning process. This method estimates energy consumption based on actual hardware measurements and models the effects of data sparsity and bit-width reduction. The algorithm prunes layers by minimizing changes in feature maps rather than filter weights, resulting in a higher compression ratio. It starts pruning from the layers with the highest energy consumption and locally fine-tunes the retained weights to restore accuracy. After pruning all layers, the entire network undergoes global fine-tuning through backpropagation. This method reduced the energy consumption of AlexNet by 3.7 times and GoogLeNet by 1.6 times, with a top-5 accuracy loss of less than 1\%. Hu et al. \cite{Hu2018ChannelPruning} proposed a novel channel pruning method based on genetic algorithms for compressing deep convolutional neural networks, significantly reducing model redundancy through a two-step optimization process, achieving performance superior to existing methods on multiple benchmark datasets. Wen et al. \cite{Wen2016LearningStructuredSparsity} proposed a structured sparse learning method to regularize the structure of DNNs, reducing computational costs, enhancing hardware-friendly sparsity, and improving classification accuracy.

\subsection{Weight Quantization}

Models are typically stored using 32-bit floating-point numbers. Weight quantization primarily involves using smaller bit widths to store model parameters, commonly using 16-bit floating-point numbers or 8-bit integers, which result in a significant reduction in the number of parameters and computational load proportional to the storage bit width; even more extreme cases involve quantizing models into binary networks \cite{Hubara2016BinarizedNeuralNetworks}, ternary weights \cite{Liu2023TWNs}, or XNOR networks \cite{Rastegari2016XNOR-Net}. For example, after simple quantization, MobileNetV1 is only 4-5 MB in size, making it easy to deploy on mobile platforms such as smartphones.

In the realm of low-bit neural network parameter representation, two quintessential quantization techniques are binary and ternary quantization. Binary quantization maps parameters to a binary set, typically consisting of -1 and +1, whereas ternary quantization allocates parameters to a ternary set, encompassing 0, +1, and -1. The application of these quantization strategies allows for the representation of parameters using a reduced bit count, thereby preserving the model's efficacy to a notable degree and facilitating the deployment of deep learning models within environments characterized by limited computational resources.

\subsection{Low-rank Approximation}

Low-rank approximation replaces a large convolution operation or fully connected operation with multiple low-dimensional operations. Common low-rank approximation methods include CP decomposition \cite{Lebedev2015CPDecomposition}, Tucker decomposition \cite{Kim2016TuckerDecomposition}, and singular value decomposition (SVD) \cite{Zhang2016AcceleratingVeryDeepCNN}. For example, if a fully connected operation of size M × N can be approximately decomposed into M × d and d × N (where d $\ll$ M, d $\ll$ N), the computational load and parameter count of this fully connected layer would be greatly reduced.

In terms of binary decomposition, Peng et al. \cite{Peng2018FilterGroupApproximation} proposed a filter group approximation-based method for low-rank decomposition, which mainly utilizes the group structure of convolution kernels at each layer to reduce parameter redundancy. The binary decomposition method introduces two tensors when applied to convolution kernels: one is a large w×h×c×d tensor, and the other is a d×n tensor. Due to the large size of the first tensor w×h×c×d and the time-consuming processing, a ternary decomposition method was proposed to achieve a more efficient decomposition. In terms of multi-way decomposition, Kim et al. \cite{Kim2016TuckerDecomposition} proposed Tucker decomposition, which further performs binary decomposition on the first tensor, resulting in a decomposition of w×1 , 1×h , and 1×1 tensors. For example, Wang et al. \cite{Wang2016AcceleratingCNN4MobileApp} proposed using the sum of several smaller tensors to approximate the original kernel tensor.

Denton et al. \cite{Denton2014LinearStructureWithinCNN} conducted an in-depth analysis of the weights of fully connected layers and convolutional layers using singular value decomposition and other matrix decomposition techniques. By appropriately truncating singular values, they successfully reduced the model's parameters and computational load while almost not compromising the network's performance. This finding has significant theoretical and practical implications for understanding the redundancy properties in deep networks and how to effectively perform network compression, providing possibilities for optimizing network design. Zhang et al. \cite{Zhang2015ApproximationNolinearCNN} explored how to further reduce the computational complexity of neural networks by approximating nonlinear activation functions. Their method innovatively reduced the demand for computational resources while maintaining high accuracy of the model. This technology breaks through traditional cognition, proving that it is still possible to effectively alleviate the computational burden of the model while maintaining high-performance standards.

\subsection{Knowledge Distillation}

Knowledge distillation was initially proposed by Buciluǎ et al. \cite{Bucilǎ2006ModelCompression}, aiming to create a compressed model by training a strong classifier with pseudo-labeled data, which can replicate the output of the original classifier. What makes knowledge distillation unique is its use of two types of networks: the teacher model and the student model. The teacher model is typically a pre-trained, large neural network model with superior performance. As shown in Figure 9, the softmax layer output of the teacher model is used as the soft target, along with the softmax layer output of the student model as the hard target, both fed into the total loss calculation to guide the training of the student model. This transfers the knowledge from the teacher model to the student model, enabling the student model to achieve performance comparable to that of the teacher model while being more compact and efficient, thus achieving the purpose of model compression.

\begin{figure}[t]
	\centering % 表示居中
	\includegraphics[scale=0.15]{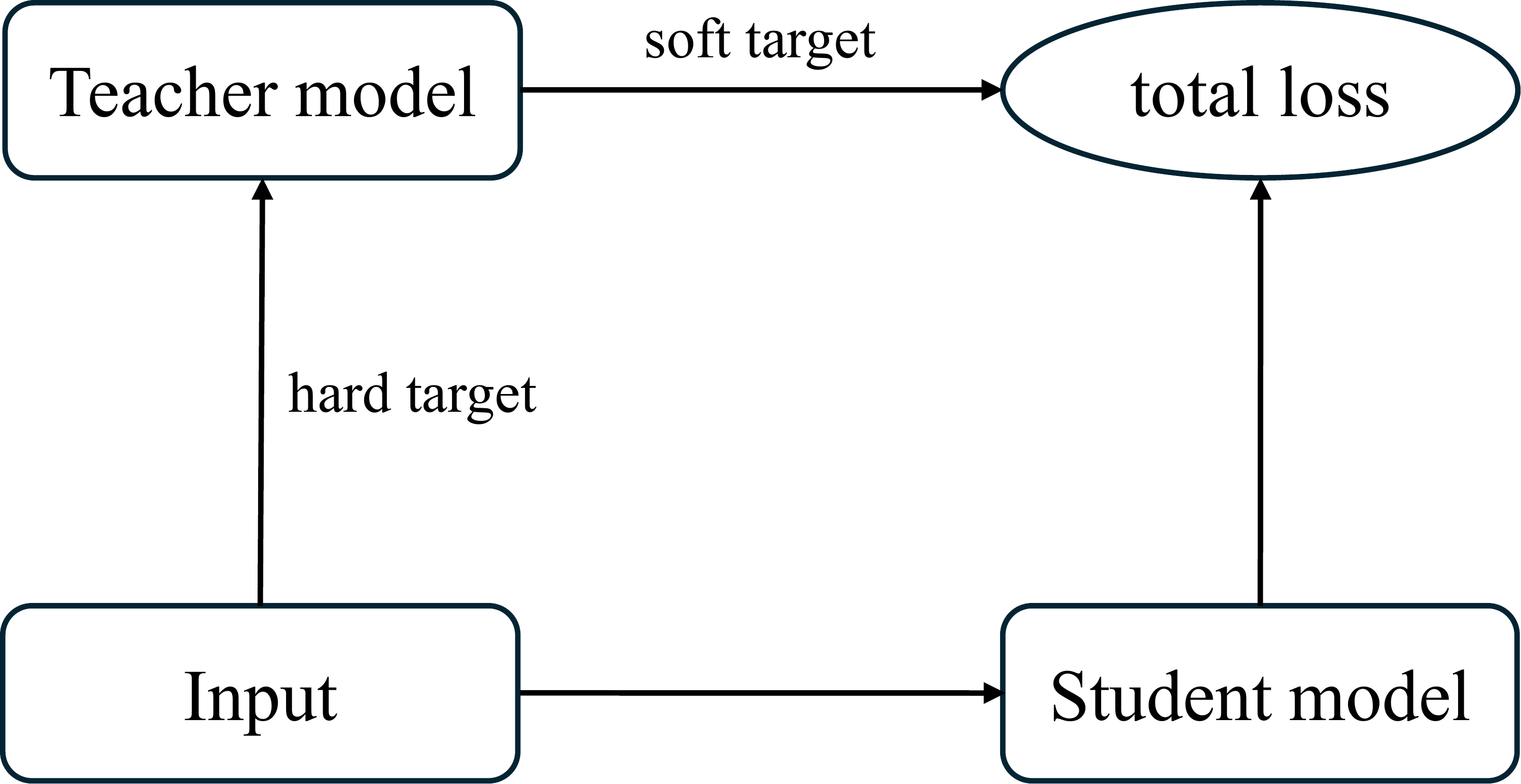}
	\caption{General process of Knowledge Distillation.}
    \label{tab: Figure 9}
\end{figure}

Hinton et al. \cite{Hinton2015DistillingTK} proposed using a high-temperature softmax to generate soft target distributions of class probabilities, and then used these soft targets to train smaller models, enabling them to be trained with less data and higher learning rates. The core idea is to train a smaller model to fit the output probability distribution (soft labels) of the teacher model, rather than directly fitting the hard labels of the teacher model. Yim et al. \cite{Yim2017GiftFromKD} proposed a novel knowledge transfer technique that achieves knowledge transfer by distilling and transferring the knowledge of a pre-trained deep neural network to another one. The distilled knowledge is defined as the flow between layers by calculating the inner product of features at two levels. Chen et al. \cite{Chen2018DarkRank} used the ranking relationships of different samples within a certain class from the teacher model as learning information to be transmitted to the student model. Czarnecki et al. \cite{Czarnecki2017Sobolev} introduced a neural network method called Sobolev training, which aims to train neural networks not only using target values but also utilizing the derivative information of the target output with respect to the input. Zhou et al. \cite{Zhou2018RocketLaunching} proposed a general training framework called "Rocket Launching" for training lightweight models. This framework utilizes a large "booster" network to supervise the learning of the lightweight network throughout the training process. The method allows the lightweight model to achieve performance close to that of deeper, more complex models by imparting the knowledge of the booster network to the lightweight network during the entire training process.

\section{Conclusion and Outlook}

The preceding research findings demonstrate that investigation of lightweight DCNNs still offers significant scope for further research. The following part in this section outlines avenues for further research and development in the field of DCNN lightweighting.

\begin{enumerate}
    \item Theoretical exploration of NAS confronts several challenges, which to a degree, restrict its practical efficacy and potential. To transcend these constraints, future investigations should concentrate on devising innovative search paradigms to enhance neural network design. This encompasses optimizing current search methodologies, innovating network architecture representation techniques, and augmenting the scope and profundity of the search space. Concurrently, it is imperative to acknowledge that the current applicability of NAS is somewhat narrow, predominantly centered on image classification. Yet, as deep learning technologies and hardware capabilities progress, an increasing number of domains are in quest of more efficient neural network configurations. Consequently, future research endeavors should prioritize broadening the application spectrum of NAS to encompass a wider array of task demands.
    \item At present, the predominant approaches to model compression each exhibit inherent limitations. Model pruning, which tends to exhibit slower convergence and necessitates a greater number of training iterations, necessitates the selection of distinct pruning strategies contingent upon the network and dataset type. Knowledge distillation, while effective, encounters limitations in its applicability and may underperform with extensive datasets. Weight quantization, which diminishes the bit-width of parameters, can diminish a model's representational capacity, notably leading to accuracy degradation in binary networks. Low-rank decomposition, conversely, exhibits heightened sensitivity to noise and outliers, potentially corrupting decomposition outcomes. Consequently, future research in model compression should concentrate on refining and amalgamating these techniques to engineer comprehensive compression frameworks that cater to specific tasks or datasets, thereby augmenting efficiency and efficacy. Moreover, aligning model compression with hardware platforms merits significant research focus. The principal challenges within model compression encompass achieving substantial compression without compromising accuracy, automating and streamlining compression processes, ensuring the compressed models' operability across diverse hardware, harmonizing theoretical frameworks with empirical results, and navigating the intricacies of multi-task and multimodal learning. The effective integration of quantization and sparsification methods to amplify compression efficacy is also pivotal. Addressing these challenges is essential for the broader and more efficient deployment of deep learning models.
    \item Interpretability represents a significant challenge for neural networks, including lightweight DCNNs. The "black box" characteristic of these models impedes the integration of model efficiency with comprehensible explanations, thereby hindering their broad implementation in practical scenarios. Consequently, it is anticipated that enhancing the interpretability of lightweight DCNNs will emerge as a prominent research focus in the forthcoming period.
\end{enumerate}

We sincerely hope that our work can provide valuable insights for researchers on lightweighting methods of DCNNs, and that they can apply these insights in their future research work.

\section*{Acknowledgements}

This research is partly supported by the Research Project of Archives Bureau, Liaoning Province, China under grant number 2024-X-016.

\bibliographystyle{cas-model2-names}

\bibliography{cas-sc-LightDCNN}

% % Here goes the biography details.

% % Here goes the biography details.

\end{document}